\documentclass[sigconf]{acmart}
\usepackage{latexsym}
\usepackage{amsmath}
\usepackage{amsthm}
\usepackage{adjustbox}
\usepackage{booktabs}
\usepackage{enumitem}
\usepackage{graphicx}
\usepackage{color}
\usepackage{multirow}
\usepackage{balance}

\newcommand{\eg}{\emph{e.g., }}

\AtBeginDocument{%
  }

\setcopyright{acmlicensed}
\copyrightyear{2024}
\acmYear{2024}
\setcopyright{acmlicensed}\acmConference[CIKM '24]{Proceedings of the 33rd
ACM International Conference on Information and Knowledge
Management}{October 21--25, 2024}{Boise, ID, USA}
\acmBooktitle{Proceedings of the 33rd ACM International Conference on
Information and Knowledge Management (CIKM '24), October 21--25, 2024,
Boise, ID, USA}
\acmDOI{10.1145/3627673.3679683}
\acmISBN{979-8-4007-0436-9/24/10}

\begin{document}

\title{Confidence-aware Self-Semantic Distillation on Knowledge Graph Embedding}

\author{Yichen Liu}
\orcid{0009-0006-3639-1666}
\affiliation{%
  \department{College of Computer Science}
  \institution{Zhejiang University}
  \city{Hangzhou}
  \country{China}
}
\email{skyminer@zju.edu.cn}

\author{Jiawei Chen}
\orcid{0000-0002-4752-2629}
\affiliation{
	 \department{College of Computer Science}
	\institution{Zhejiang University}
  \city{Hangzhou}
  \country{China}
}
\authornote{Corresponding author. }
\email{sleepyhunt@zju.edu.cn}

\author{Defang Chen}
\orcid{0000-0003-0833-7401}
\affiliation{%
  \department{College of Computer Science}
  \institution{Zhejiang University}
  \city{Hangzhou}
  \country{China}
}
\email{defchern@zju.edu.cn}

\author{Zhehui Zhou}
\orcid{0009-0009-3668-5842}
\affiliation{%
  \department{College of Computer Science}
  \institution{Zhejiang University}
  \city{Hangzhou}
  \country{China}
}
\email{zzhui@zju.edu.cn}

\author{Yan Feng}
\orcid{0000-0002-3605-5404}
\affiliation{
  \department{College of Computer Science}
  \institution{Zhejiang University}
  \city{Hangzhou}
  \country{China}
}
\email{fengyan@zju.edu.cn}

\author{Can Wang}
\orcid{0000-0002-5890-4307}
\affiliation{%
  \department{College of Computer Science}
  \institution{Zhejiang University}
  \city{Hangzhou}
  \country{China}
}
\email{wcan@zju.edu.cn}

\renewcommand{\shortauthors}{Yichen Liu et al.}

\begin{abstract}

Knowledge Graph Embedding (KGE), which projects entities and relations into continuous vector spaces, has garnered significant attention. Although high-dimensional KGE methods offer better performance, they come at the expense of significant computation and memory overheads. Decreasing embedding dimensions significantly deteriorates model performance. While several recent efforts utilize knowledge distillation or non-Euclidean representation learning to augment the effectiveness of low-dimensional KGE, they either necessitate a pre-trained high-dimensional teacher model or involve complex non-Euclidean operations, thereby incurring considerable additional computational costs. To address this, this work proposes \textit{Confidence-aware Self-Knowledge Distillation} (CSD) that learns from the model itself to enhance KGE in a low-dimensional space. Specifically, CSD extracts knowledge from embeddings in previous iterations, which would be utilized to supervise the learning of the model in the next iterations. Moreover, a specific semantic module is developed to filter reliable knowledge by estimating the confidence of previously learned embeddings. This straightforward strategy bypasses the need for time-consuming pre-training of teacher models and can be integrated into various KGE methods to improve their performance. Our comprehensive experiments on six KGE backbones and four datasets underscore the effectiveness of the proposed CSD.

\end{abstract}



\maketitle

\section{Introduction}

\begin{figure}[t]
	\centering
	\includegraphics[scale=0.42]{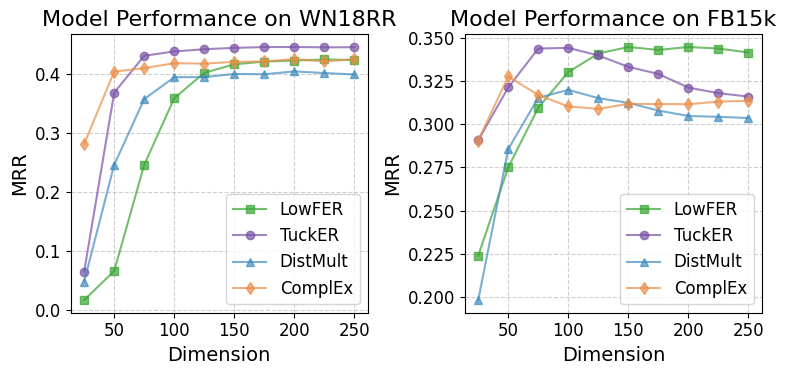}
	\caption{The comparison of model performance (MRR) with the growth of embedding dimensions of the different models on different datasets.}
	\label{marginal}
\end{figure}

Knowledge Graphs (KGs), which encapsulate real-world facts in a graph structure, have found extensive applications across various domains, ranging from event prediction \cite{EP1,EP2}, question answering \cite{QE1,QE2}, to recommender systems \cite{RS1,RS2,cui2024distillation}. As a foundational task in this area, Knowledge Graph Embedding (KGE) \cite{TransE,wu2021disenkgat} aims to represent entities and relations in KGs as continuous vectors. With the increasing scale of KGs in practice, the necessity for both accuracy and efficiency in KGE becomes increasingly critical. While recent years have witnessed a proliferation of sophisticated KGE methods, most successes hinge on high-dimensional embeddings, which inevitably incur increased memory and computational overhead. For instance, as demonstrated in Figure \ref{marginal}, representative KGE methods such as LowFER~\cite{LowFER} and DistMult~\cite{DistMult} exhibit a significant performance decline (about 96\%) in their low-dimensional versions (\eg 25 dimensions) compared to their high-dimensional (\eg 250 dimensions) counterparts, despite the low-dimensional models have 6-11 times fewer embedding layer parameters and offering 3-5 times improved inference latency.

To enhance the effectiveness of low-dimensional KGE, several recent efforts have emerged, which can be categorized into two main approaches: 1) leveraging knowledge distillation to empower a low-dimensional model with knowledge from high-dimensional teacher models \cite{chen2023unbiased,liu2023iterde,MulDE,DualDE}; and 2) embedding KGs in the non-Euclidean spaces (\eg hyperbolic or mixed-curvature spaces) to better capture the complex patterns of KGs~\cite{cao2022geometry}.  However, these strategies still face limitations: knowledge distillation requires pre-training a high-dimensional KGE model, which still demands extensive computational resources and memory; the non-Euclidean methods involve complex operations that incur additional computational costs and lack flexibility to be applied in various KGE models \cite{yao2023analogical}. Given these limitations, an intriguing research question arises: \textit{How can we achieve effective low-dimensional KGE without incurring significant additional training costs?}

To this end, this work proposes to leverage self-knowledge distillation (SKD) for both efficient and effective knowledge graph embedding with low dimensions. Unlike traditional knowledge distillation (KD), which requires a pre-trained teacher model, SKD learns from the model itself through introspection \cite{selfKD1,selfKD2,snapshot,yun2020regularizing}. Specifically, SKD utilizes information from past epochs to extract valuable knowledge to guide learning in subsequent epochs. SKD has been proven to enhance model accuracy and robustness while avoiding the substantial training overhead, and has been successfully applied in fields such as computer vision and natural language processing. However, adapting sophisticated SKD strategies to KGE presents unique challenges:

\begin{itemize}
\item \textbf{Divergent Objectives.} Recent SKD strategies are tailored for classification tasks, utilizing soft labels (\eg logits) predicted by models in earlier epochs as a medium for knowledge transfer. However, the primary goal of KGE is to generate high-quality embeddings from observed facts, whose learning procedure is typically approached as a link prediction problem. While it is possible to conceptualize KDE as a specific classification task --- where each tail node represents a distinct class --- the computation and memory demands become impractical due to the soft labels involve a vast number of entities.
\item \textbf{Inaccurate knowledge.} Given the complex structures and inherently noisy nature of KGs in practical applications, the accuracy of KGE methods may not reach the levels seen in classification tasks in NLP or CV. This suggests that self-derived knowledge may not always be reliable and can be detrimental. Empirical evidence indicates that the direct application of existing SKD methods such as CS-KD \cite{Yun_2020_CVPR} and SD \cite{yang2019snapshot} to KGE may lead to reduced performance, as detailed in Table \ref{result1}.
\end{itemize}

To overcome these challenges, we propose a novel \textit{confidence-aware self-knowledge distillation strategy} (CSD) tailored for KGE. In line with the objective of KGE, diverging from distilling knowledge from logits, CSD proposes to extract knowledge from embeddings in previous epochs, which are then utilized to guide the learning of embeddings. A specific semantic module is developed to filter reliable knowledge by estimating the confidence of previously learned embeddings. It is noteworthy that our CSD is model-agnostic. We implement our CSD across six representative KGE backbones, including non-Euclidean methods \cite{ge-etal-2023-compounding, cao2022geometry}, to demonstrate its effectiveness. To summarize, this work makes the following contributions:

\begin{itemize}
\item We advocate for leveraging self-knowledge distillation in knowledge graph embedding and highlight the associated challenges.
\item We propose an innovative iterative self-knowledge distillation method tailored for knowledge graph embedding that employs a specific semantic extraction module to distill reliable embedding knowledge from previous epochs to guide model learning.
\item We conduct extensive experiments on six backbones and four datasets to demonstrate that CSD can consistently enhance performance with limited additional computational costs.
\end{itemize}

The remainder of this paper is structured as follows. Section 2 provides a brief review of the related works. The problem definition and background are detailed in Section 3. Our proposed CSD is introduced in Section 4. Section 5 presents the experimental results and discussions. Finally, Section 6 concludes the paper and outlines potential directions for future research.

\section{Related Work}\label{rw}
\subsection{Knowledge Graph Embedding}

Numerous methods for knowledge graph embedding have been proposed, with translation-based methods developed to capture the translation property in entities. Seminal models, for instance, TransE \cite{TransE}, TransH \cite{TransH}, and TransR \cite{TransR}, project entities and relations as vectors and consider the relation as translational operations designed to transform the head entity into the tail entity. Subsequent models such as TransD \cite{TransD} and TranSparse \cite{TranSparse} extend it by adding constrain conditions for entity and relation projections, resulting in considerable performance enhancements.

Recently, there has been intensive research on a neural network model called ConvE \cite{ConvE}, which aims to capture the semantic interactions between entities and relations in knowledge graphs. Building upon this, ConvR \cite{ConvR} improves ConvE by replacing global convolution filters with relation-based filters. These filters are adaptively generated from relation embeddings, allowing ConvR to capture entity-relation interactions in the knowledge graph. Another model called HypER \cite{HypER} further extends ConvE by using a hypernetwork to generate relation-specific convolution filters through tensor factorization. This extension facilitates feature interactions between entities and relations in knowledge graphs. Additionally, AcrE \cite{AcrE} and InteractE \cite{InteractE} introduce various convolution mechanisms, such as atrous and circular convolution, to enhance entity and relation semantic interactions. JointE \cite{JointE} takes this further by jointly utilizing 1D and 2D convolutions to capture translation properties in KGs, thereby increasing interactions between entities and relations. M-DCN \cite{M-DCN} introduces a dynamic and adaptive generation of multi-scale filters for convolutions. It captures feature interactions and improves model expressiveness in knowledge graphs.

\subsection{Knowledge Distillation}

To get both performance and inference speed, knowledge distillation technology has been proposed. The initial knowledge distillation approach focused on utilizing the teacher model's prediction probabilities as soft a targets for training the student model~\cite{firstKD,wang2022semckd}. FitNet \cite{FitNet} advanced the process by using the intermediate features of the teacher model as training hints for the student model. Building on FitNet, SemCKD \cite{SemCKD} introduced semantic calibration, which employed an attention mechanism to obtain adaptive intermediate features for student training. Additionally, SimKD \cite{SimKD} achieved competitive performance for the student model simply by reusing the discriminant classifier from the teacher model. 

The pioneer in knowledge graph embedding, KD-MKB \cite{KD-MKB}, first utilized knowledge distillation for knowledge graph embedding by establishing a cooperative learning framework between a teacher model and a student model. Following this, MulDE \cite{MulDE} and DualDE \cite{DualDE} are proposed. All these papers utilized high-dimensional, powerful teacher models to facilitate the training of student models for improved performance. However, a high-capacity teacher model is not available in many cases. To address this limitation, we shifting our focus towards self-knowledge distillation strategies\cite{selfKD1,selfKD2,furlanello2018born,yun2020regularizing}. However, existing self-distillation methods predominantly cater to the field of computer vision and yet perform sub-optimally within the field of knowledge graph embedding, where they manage input data that generally assume independence from each other. However, in the Knowledge Graph Embedding (KGE) domain, the data that we input are interconnected. The current self-distillation algorithms fail to leverage this aspect of information. Consequently, to harness this component of data, we propose the Improved Self-Distillation (CSD) method. 

\begin{figure*}[pht]
	\centering
	\makebox[\textwidth][c]{\includegraphics[scale=0.4]{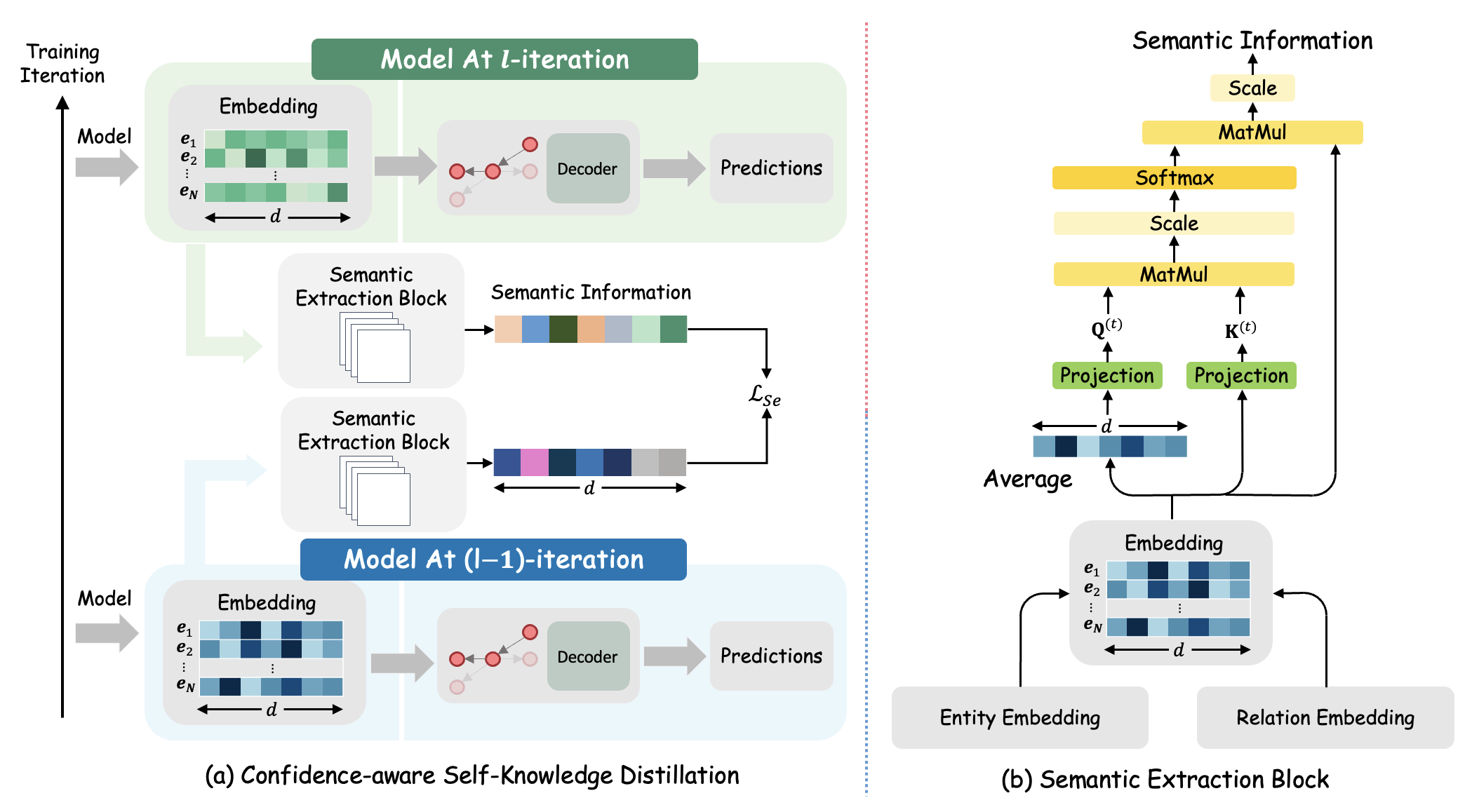}}
	\caption{The schematic of our proposed CSD. The training model is alternatively regarded as a teacher and student at the $(l-1)$-th and $l$-th iteration and distills its semantic information from embeddings through a semantic extraction block. The block filters reliable semantic knowledge by estimating the confidence of embeddings. }	\label{ISD}
\end{figure*}

\section{Preliminary}

In this section, we introduce the concepts of knowledge graph embedding and self-knowledge distillation.

\subsection{Knowledge Graph Embedding}

A knowledge graph \cite{YAGO}, denoted as $\mathcal{G} = \{ \mathcal{V},\mathcal{R}, \mathcal{T} \}$, encompasses substantial quantities of entities and relations. Herein, $\mathcal{V}$, $\mathcal{R}$ and $\mathcal{T}$ signify the sets of entities, relations, and triplets respectively. Each fact $t \in \mathcal{T}$, is presented as a triple $(h,r,t) \in \mathcal{V} \times \mathcal{R} \times \mathcal{V}$, describing there is a relationship $r \in \mathcal{R}$ from the head entity $h\in \mathcal{V}$ to the tail entity $t\in \mathcal{V}$. The target of Knowledge graph embedding (KGE) is to project each entity $v \in \mathcal{V}$ and relation $r\in \mathcal {R}$ into corresponding d-dimensional representations $\mathbf e_v, \mathbf f_r$. For convenience, we collect the embeddings of all entities and relations as matrices $\mathbf E$ and $\mathbf F$, respectively. These embeddings are subsequently utilized to facilitate downstream tasks.

To derive embeddings from Knowledge Graphs (KGs), existing Knowledge Graph Embedding (KGE) methods \cite{TuckER, DistMult, ComplEx} typically formulate the task as a specific link prediction problem. They establish a scoring function $S_{hrt} = g(\mathbf{e}_h, \mathbf{f}_r, \mathbf{e}_t)$ based on embeddings to estimate the validity of a triplet. For instance, TransE \cite{TransE} adopts $g(\mathbf{e}_h, \mathbf{f}_r, \mathbf{e}_t) = -\Vert\mathbf{e}_h + \mathbf{f}_r - \mathbf{e}_t\Vert_{1/2}$. Typically, the BCE loss can be used to optimize the KGE models as follows:
\begin{equation}
    \mathcal{L}_{BCE} = - \sum_{(h,r,t)\in \mathcal T} \log \sigma(S_{hrt})+ \sum_{(h,r,t)\in \mathcal T^-} \log (1-\sigma(S_{hrt})),
\end{equation}
where $\sigma(.)$ denotes the sigmoid function and $\mathcal T^-$ denotes a set of sampled negative instances.

\subsection{Self-Knowledge Distillation}

In order to mitigate the significant computational overhead associated with traditional knowledge distillation, self-knowledge distillation (SKD) has been explored in recent studies. The fundamental intuition behind SKD is akin to the mechanism of human introspection \cite{deng2021learning}. SKD extracts knowledge from the model's previous training epochs and subsequently utilizes this knowledge to guide its learning in the following epochs. Existing research on SKD \cite{selfKD1,selfKD2,furlanello2018born,yun2020regularizing} is primarily tailored for classification tasks, leveraging soft labels (\eg model logits) as a medium for knowledge distillation. Specifically, let $\mathbf p^{l}(x)$ be the model prediction for the sample $x$ on the $l$-th iteration. The SKD loss can be formulated as follows:
\begin{equation}
	\mathcal L_{SKD} = \sum_{x}D(\mathbf p^{l}(x),\mathbf p^{l-1}(x)),
\end{equation}
where $D(.,.)$ represents the adopted metrics, which can be implemented via KL-divergence or Huber loss, defined as:
\begin{equation}
	D_H(a, b) = 
	\begin{cases} 
	\frac{1}{2}*(a - b)^2, & |a - b| \le 1 \\
	|a -  b| - \frac{1}{2}, & | a - b| > 1.
	\end{cases}
 \label{huberloss}
\end{equation}
SKD employs a previous snapshot of the model as the supervises the learning of the current model.

\section{Methodology}\label{method}

In this section, we thoroughly discuss the proposed Confidence-aware Self-Semantic Distillation (CSD), which is tailored for low-dimensional KGE. 

As shown in Fig. \ref{ISD} (a), CSD follow sthe conventional architecture of SKD, where a KGE model alternates between the roles of teacher and student during the training process. The knowledge of the previous iteration would be utilized to supervise the learning of the current model. However, given the specific challenges in KGE, we make specific designs to adapt SKD to this task. 

Diverge from utilizing soft labels in the classification task, we distill knowledge from the model embeddings, by taking model status in $l-1$-th iteration as a teacher in $l$-th iteration. A specific semantic extraction block is developed to extract reliable knowledge from model embeddings to supervise the model learning (discussed in Section \ref{Attention Block}). 
Semantic information is dynamically and adaptively generated by the semantic extraction block, showcasing dynamic and evolution properties throughout the training process. Throughout these iterations, we leverage the teacher model to transfer semantic information to the student model by optimizing the Huber Loss between their semantic information. This optimization facilitates the student model to progressively incorporate and accumulate semantic knowledge from the past model.

\subsection{Semantic Extraction Block}\label{Attention Block}

The embedding matrices $E$ and $F$ effectively encapsulate the knowledge graph's semantic attributes based on the entities and relations. However, we find that directly distilling the embedding matrices between the teacher and student models leads to poor performance gain. This is due to the complex structures and inherently noisy nature of KGs in practical applications \cite{Paulheim2016KnowledgeGR}, where self-derived knowledge may not always be reliable and can be detrimental.

To tackle this problem, we have carefully developed an innovative semantic extraction block that dynamically generates reliable semantic knowledge. The details of this block are depicted in Fig. \ref{ISD}(b). The basic idea is to estimate the confidence of embeddings. Thus, we adaptively learn the embedding confidence, where reliability is estimated by the distance between the embedding of a certain entity (or relation) and the embedding center. Intuitively, a larger distance between an entity and the center suggests that the embedding is more likely to be an outlier. Therefore, we assign lower confidence to these embeddings.

For convenience, we primarily introduce the distillation of the entity embeddings, while relation embeddings follow a common paradigm. We first average the original entity embeddings over all entities to get their center as:

\begin{equation}
	\bar {\mathbf e} = \frac{1}{|\mathcal{V}|}\sum_{v\in \mathcal{V}} \mathbf e_v.
\end{equation}
We then calculate the confidence of embeddings by estimating their distances to the center embedding with:
\begin{equation}
c_{v}=softmax(\gamma \bar {\mathbf e} W^1_E \mathbf e_v)
\end{equation}
Here, we adopt a matrix-projected distance to measure the distance between $\bar{\mathbf e}$ and $\mathbf e_v$, where $W^1_E \in \mathbb{R}^{d\times d}$ is the product of two learnable matrices $W_{E_1} \in \mathbb{R}^{d\times k}, W_{E_2} \in \mathbb{R}^{k \times d}$ to enhance flexibility. $d$ is the dimension of $e$, and $k$ is a hyperparameter. We mapped $\bar{\mathbf e}$ and $\mathbf e_v$ into k-dimension and computed their distance via dot production. $\gamma$ is introduced to control the scaling, while $softmax(.)$ is utilized for normalization. Finally, the extracted semantic knowledge can be integrated as follows:
\begin{equation}
I_E=\sum_{v\in \mathcal{V}}{c_{v}\mathbf e_vW^2_E}.
\end{equation}
Here, we aggregate the information from each entity to generate the final extracted knowledge for knowledge distillation, where the contribution of each entity is discounted by its confidence. Also, we introduce another parameter $W^2_E$to map embeddings into another space to facilitate the distillation. This approach significantly downweights the impact of unreliable and low-quality embeddings. Similarly, the extracted knowledge of relation embeddings $I_F$ can be calculated in a similar way. 

A Huber function is utilized to supervise the learning of the model:
\begin{equation}
    \mathcal{L}_{se} = Huber(I_E^l, I_E^{l-1}) + Huber(I_F^l, I_F^{l-1}),
\end{equation}
where $I_E^l$ and $I_E^{l-1}$ denote the extracted knowledge of entity embeddings in the $l-th$ and $(l-1)-th$ iterations. The introduced distillation loss provides a new signal that guides the current model to align the semantic knowledge extracted from the model's past epochs, thereby leading to performance gains.

\subsection{Training Objectives}

Given that the scores $S_{hrt}$reflect the model's prediction on the validity of a fact, which may also contain some valuable signals, we also introduce an additional loss to distill the knowledge from $S_{hrt}$:
\begin{equation}
	\mathcal{L}_{sc} = \sum_{(h,r,t)\in \mathcal{T}}Huber(S^t_{hrt}, S^s_{hrt}),
	\label{lossSD}
\end{equation}
where $S^t_{hrt}$ denotes the score from the model's past epochs, and $S^s_{hrt}$ denotes the scores predicted by the current model.

Finally, the overall loss can be formulated as follows:
\begin{equation}
	\mathcal{L} = (1-\beta)\mathcal{L}_{BCE} + \beta\mathcal{L}_{se} + \lambda\mathcal{L}_{so},
	\label{loss}
\end{equation}
\noindent
where $\beta$ and $\lambda$ balance the contributions from each component.

\section{Experiments}\label{experiments}

In this section, we undertake an extensive array of experiments aimed at addressing these key questions:

\begin{itemize} 
    \item \textbf{RQ1:} How does CSD compare with existing self-distillation methods when applied to KGE models? 
    \item \textbf{RQ2:} How does CSD compare with well-established distillation techniques frequently utilized in knowledge graph embedding?
    \item \textbf{RQ3:} How effectively does CSD operate across varying dimensions? 
    \item \textbf{RQ4:} How do the model's hyperparameters influence its performance? 
    \item \textbf{RQ5:} What is the impact of these partial modules - iterative distillation, periodic snapshot distillation, and confidence-aware semantic distillation - in enhancing the model's efficiency?
\end{itemize}

\subsection{Experiments Settings}

\subsubsection{Datasets}

Comprehensive link prediction experiments are performed on three standard benchmark datasets: YAGO3-10~\cite{mahdisoltani2013yago3}, FB15K-237~\cite{FB15k237}, and WN18RR \cite{ConvE}. These datasets have been extensively utilized in knowledge graph embedding studies. Note that datasets FB15K-237 and WN18RR are subsets of FB15k~\cite{TransE} and WN18~\cite{TransE}, respectively. FB15k and WN18 exhibit test data leakage, where numerous inverse relations from the testing sets can be conveniently discovered in the training sets \cite{toutanova2015representing}. FB15K-237 and WN18RR datasets were generated by eliminating inverse relations to rectify this problem. We therefore do not test our method on FB15k and WN18 datasets. Detailed statistics of these datasets are provided in Table \ref{datasets}.

\begin{table}[htbp]
	\setlength{\tabcolsep}{2mm}{}
	\centering
	\normalsize
	\caption{The detailed statistics of three standard benchmark datasets. $|\mathit{R}|$ and $|\mathcal{V}|$ denotes the number of relations and entities in the corresponding dataset. $|\mathcal{T}_{train}|$ denotes the number of triplets in training set. }\label{datasets}
	\begin{tabular}[t]{cccc}
		\toprule
		Dataset&FB15K-237& WN18RR & YAGO3-10 \\
		\midrule
		$|\mathit{R}|$ & 237 & 11 & 37 \\
		$|\mathcal{V}|$ & 14,541 & 40,943 & 123,182 \\
		$|\mathcal{T}_{train}|$ & 272,115 & 86,835 & 1,079,040\\
		$|\mathcal{T}_{valid}|$ & 17,535 & 3,034 & 5,000 \\
		$|\mathcal{T}_{test}|$ & 20,446 & 3,134 & 5,000 \\
		\bottomrule
	\end{tabular}	
\end{table}

\begin{table*}[htbp]
	\setlength{\tabcolsep}{1mm}{}
	\centering
	\normalsize
	\caption{Comparison results of link prediction on YAGO3-10, WN18RR and FB15K-237 datasets with 25 dimensions. The $Gain$ is the improvement of the CSD ratio relative to the baseline. The Baseline refers to the results obtained from directly training the models without applying any distillation methods. The - indicates that the method cannot be applied in the corresponding model in our hardware due to being out of memory.}\label{result1}
	\resizebox{\textwidth}{!}{
        \begin{adjustbox}{width=\textwidth,center}
	\begin{tabular}[t]{cccccccccccccccccccc}
		\toprule
		\multirow{3}*{Dataset} & \multirow{3}*{Method}  & \multicolumn{3}{c}{DistMult} & \multicolumn{3}{c}{ComplEx} & \multicolumn{3}{c}{LowFER} & \multicolumn{3}{c}{TuckER} & \multicolumn{3}{c}{GIE} & \multicolumn{3}{c}{CompoundE} \\
		\cmidrule(r){3-5}  \cmidrule(r){6-8} \cmidrule(r){9-11} \cmidrule(r){12-14}  \cmidrule(r){15-17} \cmidrule(r){18-20}
		 &  & MRR & H@10 & H@1 & MRR & H@10 & H@1 & MRR & H@10 & H@1 & MRR & H@10 & H@1 & MRR & H@10 & H@1 & MRR & H@10 & H@1 \\
		\midrule
        \multirow{7}*{YAGO3-10} & Baseline & .1043 & .1757 & .0657 & .1868 & .3136 & .1214 & .0510 & .0702 & .0376 & .0782 & .1217 & .0513 & .4546 & .6040 & .3758 & .1983 & .3439 & \textbf{.1263} \\
							  & CS-KD & .0641 & .0970 & .0447 & .1284 & .2174 & .0814 & .0376 & .0511 & .0278 & .0612 & .1022 & .0377 & .1626 & .2824 & .1005 & - & - & - \\
							  & SD & .1082 & .1835 & .0677 & .2116 & .3490 & .1403 & .0686 & .1009 & \textbf{.0472} & .1031 & .1744 & .0650 & .4106 & .5570 & .3326 & .1694 & .3049 & .1019 \\
							  \cmidrule{2-20}
							  & \textbf{CSD} & \textbf{.1134} & \textbf{.1903} & \textbf{.0724} & \textbf{.2139} & \textbf{.3553} & \textbf{.1411} & \textbf{.0702} & \textbf{.1064} & .0470 & \textbf{.1117} & \textbf{.1874} & \textbf{.0708} & \textbf{.5074} & \textbf{.6423} & \textbf{.4320} & \textbf{.1989} & \textbf{.3489} & .1253 \\
							  & $\mathbf{Gain_{Baseline}}\uparrow$ & \textbf{8.7\%} & \textbf{8.3\%} & \textbf{10.2\%} & \textbf{14.5\%} & \textbf{13.3\%} & \textbf{16.2\%} & \textbf{37.6\%} & \textbf{51.6\%} & 25.0\% & \textbf{42.8\%} & \textbf{54.0\%} & \textbf{38.0\%} & \textbf{11.6\%} & \textbf{6.3\%} & \textbf{15.0\%} & \textbf{0.3\%} & \textbf{1.5\%} & -0.8\%  \\
                                & $\mathbf{Gain_{SD}}\uparrow$ & \textbf{4.8\%} & \textbf{3.7\%} & \textbf{6.9\%} & \textbf{1.1\%} & \textbf{1.8\%} & \textbf{0.6\%} & \textbf{2.3\%} & \textbf{5.5\%} & -0.4\% & \textbf{8.3\%} & \textbf{7.5\%} & \textbf{8.9\%} & \textbf{23.6\%} & \textbf{15.3\%} & \textbf{29.9\%} & \textbf{17.4\%} & \textbf{14.4\%} & 23.0\% \\
        \midrule
		\multirow{8}*{WN18RR} & Baseline & .0474 & .0811 & .0278 & .2806 & .3992 & .2179 & .0160 & .0282 & .0078 & .0644 & .1099 & .0393 & .4149 & .4525 & .3955 & .3868 & .4833 & .3322 \\
							  & CS-KD & .0255 & .0431 & .0148 & .1242 & .1723 & .0968 & .0092 & .0180 & .0040 & .0266 & .0487 & .0144 & .3232 & .3907 & .2865 & - & - & - \\
							  & LWR & .0436 & .0767 & .0255 & .2287 & .3445 & .1685 & .0129 & .0233 & .0065 & .0599 & .1069 & .0346 & - & - & - & - & - & - \\ 
							  & SD & .0774 & .1241 & .0520 & \textbf{.3586} & .4778 & \textbf{.2910} & .0305 & .0539 & .0171 & .1840 & .2824 & .1337 & .4231 & .4566 & .4048 & .3859 & .4820 & .3301 \\
							  \cmidrule{2-20}
							  & \textbf{CSD} & \textbf{.1160} & \textbf{.1878} & \textbf{.0782} & .3581 & \textbf{.4789} & .2883 & \textbf{.0607} & \textbf{.1106} & \textbf{.0343} & \textbf{.2230} & \textbf{.3325} & \textbf{.1658} & \textbf{.4328} & \textbf{.4856} & \textbf{.4060} & \textbf{.3997} & \textbf{.4946} & \textbf{.3454} \\
							  & $\mathbf{Gain_{Baseline}}\uparrow$ & \textbf{144.7\%} & \textbf{131.6\%} & \textbf{181.3\%} & 27.6\% & \textbf{20.0\%} & 32.3\% & \textbf{279.4\%} & \textbf{292.2\%} & \textbf{339.7\%} & \textbf{246.3\%} & \textbf{202.5\%} & \textbf{321.9\%} & \textbf{4.3\%} & \textbf{7.3\%} & \textbf{2.7\%} & \textbf{3.3\%} & \textbf{2.3\%} & \textbf{4.0\%} \\
                                & $\mathbf{Gain_{SD}}\uparrow$ & \textbf{49.9\%} & \textbf{51.3\%} & \textbf{50.4\%} & -0.1\% & \textbf{0.2\%} & -0.9\% & \textbf{99.0\%} & \textbf{105.2\%} & \textbf{100.6\%} & \textbf{21.2\%} & \textbf{17.7\%} & \textbf{24.0\%} & \textbf{2.3\%} & \textbf{6.4\%} & \textbf{0.3\%} & \textbf{3.6\%} & \textbf{2.6\%} & \textbf{4.6\%} \\

		\midrule
		\multirow{8}*{FB15K-237} & Baseline & .1983 & .3039 & .1438 & .2906 & .4462 & .2110 & .2240 & .3452 & .1616 & .2680 & .4125 & .1940 & .3301 & .5059 & .2420 & .2847 & .4546 & .1988 \\
							  & CS-KD & .1396 & .2229 & .0997 & .2400 & .3737 & .1710 & .1238 & .2263 & .0703 & .1741 & .3082 & .1058 & .2065 & .3767 & .1222 & - & - & - \\
							  & LWR & .1918 & .2937 & .1384 & .2849 & .4397 & .2057 & .2124 & .3263 & .1533 & .2564 & .3958 & .1849 & - & - & - & - & - & - \\ 
							  & SD & .2309 & .3569 & .1649 & .3083 & .4740 & .2244 & .2352 & .3633 & .1689 & .2878 & .4407 & .2102 & .3291 & .5013 & .2426 & .2829 & .4534 & .1975 \\
							  \cmidrule{2-20}
							  & \textbf{CSD} & \textbf{.2545} & \textbf{.3923} & \textbf{.1837} & \textbf{.3113} & \textbf{.4795} & \textbf{.2257} & \textbf{.2394} & \textbf{.3703} & \textbf{.1719} & \textbf{.2927} & \textbf{.4464} & \textbf{.2140} & \textbf{.3316} & \textbf{.5092} & \textbf{.2436} & \textbf{.2867} & \textbf{.4562} & \textbf{.2025} \\
							  & $\mathbf{Gain_{Baseline}}\uparrow$ & \textbf{28.3\%} & \textbf{29.1\%} & \textbf{27.7\%} & \textbf{7.1\%} & \textbf{7.5\%} & \textbf{7.0\%} & \textbf{6.9\%} & \textbf{7.3\%} & \textbf{6.4\%} & \textbf{9.2\%} & \textbf{8.2\%} & \textbf{10.3\%} & \textbf{0.5\%} & \textbf{0.7\%} & \textbf{0.7\%} & \textbf{0.7\%} & \textbf{0.4\%} & \textbf{1.9\%} \\
                                & $\mathbf{Gain_{SD}}\uparrow$ & \textbf{10.2\%} & \textbf{9.9\%} & \textbf{11.4\%} & \textbf{1.0\%} & \textbf{1.2\%} & \textbf{0.6\%} & \textbf{1.8\%} & \textbf{1.9\%} & \textbf{1.8\%} & \textbf{1.7\%} & \textbf{1.3\%} & \textbf{1.8\%} & \textbf{0.8\%} & \textbf{1.6\%} & \textbf{0.4\%} & \textbf{1.3\%} & \textbf{0.6\%} & \textbf{2.5\%} \\
		\bottomrule
	\end{tabular}
        \end{adjustbox} 
	}
\end{table*}

\subsubsection{Baselines}

We select some typical KGE models as baselines: (1) DistMult~\cite{DistMult}: it exploits bilinear objective to project the entities and relations into vectors and matrices, respectively. (2) ComplEx \cite{ComplEx}: it exploits bilinear objective to project the entities and relations into vectors and matrices, respectively. (3) TuckER \cite{TuckER}: it leverages Tucker decomposition to obtain the entity and relation representations. (4) LowFER \cite{LowFER}: it is an extension of TuckER by employing multi-modal factorized bilinear pooling to improve the model expressiveness of entity and relation embeddings. (5) GIE \cite{cao2022geometry}: it is a cutting-edge method for embedding knowledge graphs. (6) CompoundE \cite{ge-etal-2023-compounding}: it is a pioneering knowledge graph embedding model that maximizes the utility of geometric transformations, namely translation, rotation, and scaling. 

We compare CSD with the standard training produced and recent self-distillation methods: 

\begin{itemize}
	\item \textbf{CS-KD} \cite{Yun_2020_CVPR}: CS-KD assumes that data within the same class share some common information. Consequently, it conducts distillation between different samples of the same class. To expedite training KGE models, multiple entities were amalgamated for distillation in our experiments.
        \item \textbf{LWR} \cite{deng2021learning}: During the training process of a single model, LWR records the logit of each sample in the earlier training process, and uses the recorded logit for distillation in subsequent training.
        \item \textbf{SD} \cite{yang2019snapshot}: SD uses a periodic learning rate strategy to generate the teacher model and utilize its prediction logits for knowledge distillation.
\end{itemize}

\subsubsection{Metrics}

Following prior work \cite{liu2023iterde}, we measure the performance of knowledge graph embedding models on link prediction tasks by Mean Reciprocal Ranks (MRR), H@10 and H@1 metrics.

\subsubsection{Implementation Details}\label{implement}

We implement all models with the proposed CSD strategy in PyTorch. All experiments are conducted on an NVIDIA GeForce GTX 4090 GPU(24GB). For general link prediction experiments, we utilize the Adam \cite{Adam} optimizer. Batch normalization \cite{BN} and dropout \cite{dropout} are used to mitigate over-fitting. To investigate the model performance in the low-dimension space, we set the embedding dimensions to 25 and 100. Other detailed parameter settings are approximately identical to their original papers. All results of * are retrieved from the paper of DualDE~\cite{DualDE} or IterDE~\cite{liu2023iterde}. We trained 500 epochs for each knowledge graph model. For the CSD strategy, we set $L$ to $125$, select $k = 50, \gamma=0.25$ for 25 dimensions and $k=200, \gamma=0.1$ for 100 dimensions. We search the value of $\beta$ and $\lambda$ in the interval $(0, 1)$. For every sub-training process, $\beta$ will be reset to the initial setting value and decay over training epochs to take advantage of the temporal attentiveness. The value of $\beta$ multiplied by $(1-ep/EP)$ at the beginning of $ep$-th epoch in this sub-training process, where $EP$ is the total number of epochs for each sub-training process.

\subsection{Link Prediction Results (RQ1 and RQ2)}\label{link prediction}

We perform link prediction experiments for all baselines combined with our method. The experimental results of 50 and 100 dimensions are shown in Tables \ref{result1} and \ref{result2}. The best results are in bold. Overall, the baseline model combined with our method outperforms all original baselines on these datasets. Specifically, the LowFER model combined with CSD with 25 dimensions has significant performance gains on the WN18RR dataset with improvements of 279.4\% (from 0.0160 to 0.0607), 292.2\% (from 0.0282 to 0.1106) and 339.7\% (from 0.0078 to 0.0343) in MRR, H@10 and H@1, respectively. 

\noindent
\textbf{Comparing CSD with CS-KD and LWR}. Our CSD surpasses both CS-KD and LWR across all backbones and datasets, thereby substantiating CSD's efficacy. It merits mentioning that employing the comparative method, LWR, on the YAGO3-10 dataset proves challenging due to the necessity of recording the model's output logits for each sample, resulting in considerable memory consumption. And the comparative methods CS-KD and LWR do not surpass the baseline performance. CS-KD is primarily designed for image classification tasks involving limited classes and encounters challenges in the link prediction task. The large number of entities in this task significantly exceeds the class quantity in image classification tasks, resulting in insufficient samples for CS-KD to acquire class-specific knowledge, thereby leading to performance degradation. LWR's approach to directly distill knowledge from a weak prediction to the model proves effective in image classification tasks, owed primarily to the dark knowledge contained in the weak prediction distribution. Conversely, in the link prediction task, the dark knowledge of a weak prediction distribution declines due to the immense number of entities involved. Consequently, the detrimental impact of a weak prediction distribution outweighs its benefit, some erroneous prediction distribution for knowledge distillation may cause the student model to undergo performance degradation.

\noindent
\textbf{Comparing CSD with SD}. Our proposed CSD outperforms SD on almost every backbone and dataset. SD, akin to our proposed CSD, distills knowledge of the teacher model obtained through the periodic learning rate strategy. Despite SD yielding satisfactory results on certain models, some instances revealed that the Mean Reciprocal Rank (MRR) performance was inferior to the Baseline. The inclusion of a semantic extraction block facilitates the student model in acquiring semantic knowledge based on the last iteration. This Additional advantage allows CSD to surpass SD in performance across the majority of models and datasets.

\noindent
\textbf{Comparing CSD with Baseline}. The proposed CSD exemplifies superior performance compared to the Baseline across all backbones and datasets with 25 dimensions and outshines nearly all backbones and datasets with 100 dimensions. Apart from evaluating the Mean Reciprocal Rank (MRR), a comparative analysis was also conducted on the training time for the CSD and Baseline. As demonstrable in Table \ref{time_comparison}, CSD necessitates a longer training duration than the Baseline due to the semantic extraction block, indicating a trade-off between MRR performance and training time. However, it is noteworthy that during model inference, the semantic extraction block is non-participatory in computations. Hence, models trained using the CSD method match the Baseline's inference speeds.

\begin{table*}[htbp]
	\setlength{\tabcolsep}{1mm}{}
	\centering
	\normalsize
	\caption{Comparison results of link prediction on YAGO3-10, WN18RR and FB15K-237 datasets with 100 dimensions.}\label{result2}
	\resizebox{\textwidth}{!}{
        \begin{adjustbox}{width=\textwidth,center}
	\begin{tabular}[t]{cccccccccccccccccccc}
		\toprule
		\multirow{3}*{Dataset} & \multirow{3}*{Method}  & \multicolumn{3}{c}{DistMult} & \multicolumn{3}{c}{ComplEx} & \multicolumn{3}{c}{LowFER} & \multicolumn{3}{c}{TuckER} & \multicolumn{3}{c}{GIE} & \multicolumn{3}{c}{CompoundE} \\
		\cmidrule(r){3-5}  \cmidrule(r){6-8} \cmidrule(r){9-11} \cmidrule(r){12-14}  \cmidrule(r){15-17} \cmidrule(r){18-20}
		 &  & MRR & H@10 & H@1 & MRR & H@10 & H@1 & MRR & H@10 & H@1 & MRR & H@10 & H@1 & MRR & H@10 & H@1 & MRR & H@10 & H@1 \\
		\midrule
        \multirow{7}*{YAGO3-10} & Baseline & .3605 & .5224 & .2741 & .5517 & .6864 & .4764 & .1983 & .3307 & .1298 & .3338 & .5055 & .2466 & .5539 & .6848 & .4822 & .3748 & .5717 & .2715 \\
							  & CS-KD & .2346 & .3660 & .1679 & .3389 & .4741 & .2676 & .1652 & .2835 & .1052 & .2573 & .4024 & .1836 & .2896 & .4416 & .2091 & - & - & - \\
							  & SD & .4432 & .6059 & \textbf{.3567} & .5634 & .6948 & .4901 & .2746 & .4394 & .1914 & .3515 & .5281 & .2594 & .5512 & .6771 & .4796 & .3711 & .5665 & .2687 \\
							  \cmidrule{2-20}
							  & \textbf{CSD} & \textbf{.4455} & \textbf{.6111} & \textbf{.3567} & \textbf{.5670} & \textbf{.6958} & \textbf{.4942} & \textbf{.2914} & \textbf{.4618} & \textbf{.2060} & \textbf{.3533} & \textbf{.5298} & \textbf{.2626} & \textbf{.5599} & \textbf{.6894} & \textbf{.4883} & \textbf{.3969} & \textbf{.5796} & \textbf{.2996} \\
							  & $\mathbf{Gain_{Baseline}}\uparrow$ & \textbf{23.6\%} & \textbf{17.0\%} & \textbf{30.1\%} & \textbf{2.8\%} & \textbf{1.4\%} & \textbf{3.7\%} & \textbf{46.9\%} & \textbf{39.6\%} & \textbf{58.7\%} & \textbf{5.8\%} & \textbf{4.8\%} & \textbf{6.5\%} & \textbf{1.1\%} & \textbf{0.7\%} & \textbf{1.3\%} & \textbf{5.9\%} & \textbf{1.4\%} & \textbf{10.3\%} \\
                                & $\mathbf{Gain_{SD}}\uparrow$ & \textbf{0.5\%} & \textbf{0.9\%} & \textbf{0.0\%} & \textbf{0.6\%} & \textbf{0.1\%} & \textbf{0.8\%} & \textbf{6.1\%} & \textbf{5.1\%} & \textbf{7.6\%} & \textbf{0.5\%} & \textbf{0.3\%} & \textbf{1.2\%} & \textbf{1.6\%} & \textbf{1.8\%} & \textbf{1.8\%} & \textbf{7.0\%} & \textbf{2.3\%} & \textbf{11.5\%} \\
        \midrule
		\multirow{8}*{WN18RR} & Baseline & .3939 & .4620 & .3607 & .4178 & .4905 & .3811 & .3584 & .4140 & .3275 & .4337 & .4708 & .4140 & .4424 & .4895 & \textbf{.4191} & .4443 & .5233 & .4040 \\
							  & CS-KD & .3543 & .4434 & .3025 & .4056 & .4757 & .3711 & .2919 & .3668 & .2526 & .4310 & .4654 & .4124 & .3277 & .4190 & .2770 & - & - & - \\
							  & LWR & .3951 & .4588 & .3630 & .4251 & .4983 & .3906 & .3567 & .4162 & .3231 & .4315 & .4724 & .4092 & - & - & - & - & - & - \\ 
							  & SD & .4133 & .4912 & .3767 & .4365 & \textbf{.5171} & .3984 & .4449 & .4927 & .4201 & .4503 & .4978 & .4253 & .4324 & .4754 & .4094 & .4598 & .5416 & .4185 \\
							  \cmidrule{2-20}
							  & \textbf{CSD} & \textbf{.4141} & \textbf{.4927} & \textbf{.3786} & \textbf{.4389} & .5147 & \textbf{.4025} & \textbf{.4612} & \textbf{.5127} & \textbf{.4324} & \textbf{.4542} & \textbf{.4995} & \textbf{.4298} & \textbf{.4465} & \textbf{.5115} & .4148 & \textbf{.4687} & \textbf{.5445} & \textbf{.4281} \\
							  & $\mathbf{Gain_{Baseline}}\uparrow$ &\textbf{5.1\%} & \textbf{6.6\%} & \textbf{5.0\%} & \textbf{5.1\%} & 4.9\% & \textbf{5.6\%} & \textbf{28.7\%} & \textbf{23.8\%} & \textbf{32.0\%} & \textbf{4.7\%} & \textbf{6.1\%} & \textbf{3.8\%} & \textbf{0.9\%} & \textbf{4.5\%} & -1.0\% & \textbf{5.5\%} & \textbf{4.1\%} & \textbf{6.0\%} \\
                                & $\mathbf{Gain_{SD}}\uparrow$ & \textbf{0.2\%} & \textbf{0.3\%} & \textbf{0.5\%} & \textbf{0.5\%} & -0.5\% & \textbf{1.0\%} & \textbf{3.7\%} & \textbf{4.1\%} & \textbf{2.9\%} & \textbf{0.9\%} & \textbf{0.3\%} & \textbf{1.1\%} & \textbf{3.3\%} & \textbf{7.6\%} & 1.3\% & \textbf{1.9\%} & \textbf{0.5\%} & \textbf{2.3\%} \\
		\midrule
		\multirow{8}*{FB15K-237} & Baseline & .3199 & .4967 & .2318 & .3103 & .4873 & .2235 & .3302 & .5035 & .2435 & .3442 & .5270 & .2529 & .3487 & \textbf{.5366} & .2563 & .3147 & .4949 & .2245 \\
								 & CS-KD & .3157 & .4891 & .2286 & .3122 & .4861 & .2248 & .3146 & .4797 & .2321 & .3382 & .5199 & .2472 & .2405 & .4293 & .1480 & - & - & - \\
								 & LWR & .3175 & .4911 & .2296 & .3108 & .4876 & .2229 & .3251 & .4963 & .2404 & .3386 & .5223 & .2465 & - & - & - & - & - & - \\
								 & SD & .3274 & .5078 & \textbf{.2374} & .3164 & .4942 & .2292 & .3476 & .5254 & .2578 & .3457 & \textbf{.5301} & .2530 & .3502 & .5350 & \textbf{.2582} & .3168 & .4978 & .2264 \\
								 \cmidrule{2-20}
								 & \textbf{CSD} & \textbf{.3275} & \textbf{.5079} & .2366 & \textbf{.3185} & \textbf{.4965} & \textbf{.2302} & \textbf{.3553} & \textbf{.5370} & \textbf{.2637} & \textbf{.3485} & .5286 & \textbf{.2582} & \textbf{.3512} & .5357 & .2581 & \textbf{.3194} & \textbf{.4993} & \textbf{.2301} \\
								 & $\mathbf{Gain_{Baseline}}\uparrow$ & \textbf{2.4\%} & \textbf{2.3\%} & 2.1\% & \textbf{2.6\%} & \textbf{1.9\%} & \textbf{3.0\%} & \textbf{7.6\%} & \textbf{6.7\%} & \textbf{8.3\%} & \textbf{1.2\%} & 0.3\% & \textbf{2.1\%} & \textbf{0.7\%} & -0.2\% & 0.7\% & \textbf{1.5\%} & \textbf{0.9\%} & \textbf{2.4\%} \\
                                 & $\mathbf{Gain_{SD}}\uparrow$ & 0.0\% & 0.0\% & -0.3\% & \textbf{0.7\%} & \textbf{0.5\%} & \textbf{0.4\%} & \textbf{2.2\%} & \textbf{2.2\%} & \textbf{2.3\%} & \textbf{0.8\%} & -0.3\% & \textbf{2.1\%} & \textbf{0.3\%} & 0.1\% & 0.0\% & \textbf{0.8\%} & \textbf{0.3\%} & \textbf{1.6\%} \\
		\bottomrule
	\end{tabular}
        \end{adjustbox} 
	}
\end{table*}

\begin{table}[htbp]
	\setlength{\tabcolsep}{0.8mm}{}
	\centering
	\normalsize
	\caption{Comparison of training time (minutes) between CSD and Baseline with 25 dimensions.}\label{time_comparison}
	\begin{tabular}[t]{ccccccc}
		\toprule
	& & DistMult & ComplEx & LowFER & TuckER & GIE \\
		\midrule
            \multirow{2}*{WN18RR} & Baseline & 151 & 154 & 152 & 152 & 13 \\
        & CSD & 156 & 158 & 159 & 156 & 21 \\
	\cmidrule{1-7}
        \multirow{2}*{FB15k-237} & Baseline & 51 & 59 & 50 & 51 & 44 \\
        & CSD & 58 & 75 & 63 & 57 & 57 \\
	\bottomrule
	\end{tabular}
\end{table}

\begin{table}[htbp]
	\setlength{\tabcolsep}{1mm}{}
	\centering
	\normalsize
	\caption{Link prediction results on the ComplEx model.}
    \label{t_comparison}
	\begin{tabular}[t]{cccccccc}
		\toprule \multirow{3}*{\textbf{Dim}} & 
		 \multirow{3}*{\textbf{Method}} & \multicolumn{3}{c}{WN18RR} & \multicolumn{3}{c}{FB15K-237} \\
		\cmidrule(r){3-5}  \cmidrule(r){6-8} 
	& & MRR & H@10 & H@1 & MRR & H@10 & H@1 \\
		\midrule
         512 & Tea.* & .433 & .515 & .387 & .298 & .472 & .213 \\
            \midrule
            \multirow{6}*{64} & BKD* & .377 & .475  & .330 & .258 & .422 & .187 \\
        & RKD* & .396 & .498 & .343 & .295 & .470 & .212\\
        & TA* & .407 & .494 & .362 & .259 & .423 & .177 \\
        & DualDE* & .419 & .507 & .379 & .303 & .478 & .218 \\
        & IterDE & .398 & .452  & .368 & .258 & .426 & .174 \\
	\cmidrule{2-8}
        & \textbf{CSD} & \textbf{430} & \textbf{.513} & \textbf{.390}  & \textbf{.324} & \textbf{.506} & \textbf{.235} \\
        \cmidrule{1-8}
        \multirow{6}*{32} & BKD* & .343 & .406  & .302 & .239 & .394 & .161 \\
        & RKD* & .368 & .456 & .322 & .270 & .440 & .185 \\
        & TA* & .372 & .464 & .315 & .259 & .423 & .177 \\
        & DualDE* & .397 & .473 & .352 & .274 & .444 & .189 \\
        & IterDE* & .400 & .484 & .374 & .277 & .454 & .192 \\
	\cmidrule{2-8}
        & \textbf{CSD} & \textbf{.410} & \textbf{.496} & \textbf{.362}  & \textbf{.323} & \textbf{.496} & \textbf{.236} \\
	\bottomrule
	\end{tabular}
\end{table}

We executed a comparative analysis of our CSD method with existing and well-established distillation techniques frequently utilized in low-dimensional knowledge graph embedding.  The methods include: (1) \textbf{BKD}\cite{firstKD}: The first knowledge distillation method. (2) \textbf{RKD}\cite{park2019relational}: This distillation technique calculates the distillation loss by evaluating the distance and angle of the embedding vector between the teacher and the student models. (3) \textbf{TA}\cite{mirzadeh2020improved}: This method chooses an intermediary model between the teacher and the student models for the distillation. (4)\textbf{DualDE} \cite{DualDE}: This implements a two-stage distillation integrated with the structural distillation technique. (5) \textbf{IterDE} \cite{liu2023iterde}: This technique utilizes a multi-stage distillation strategy. The experiment results are shown in Table \ref{t_comparison}.

\noindent \textbf{Comparing CSD with the teacher model.} Our technique brings the performance of models having 64 and 32 dimensions either close to or even superior to that of the 512-dimensional teacher model. This once again proves that our method can effectively improve the expression ability of low-dimensional models. 

\noindent \textbf{Comparing CSD with existing distillation strategies.} CSD surpasses almost all current distillation approaches. Notably, our technique does not employ the teacher model, yet performance still surpasses all techniques that utilize a teacher model for distillation.

\begin{figure*}[!h]
	\centering
	\makebox[\textwidth][c]{\includegraphics[scale=0.44]{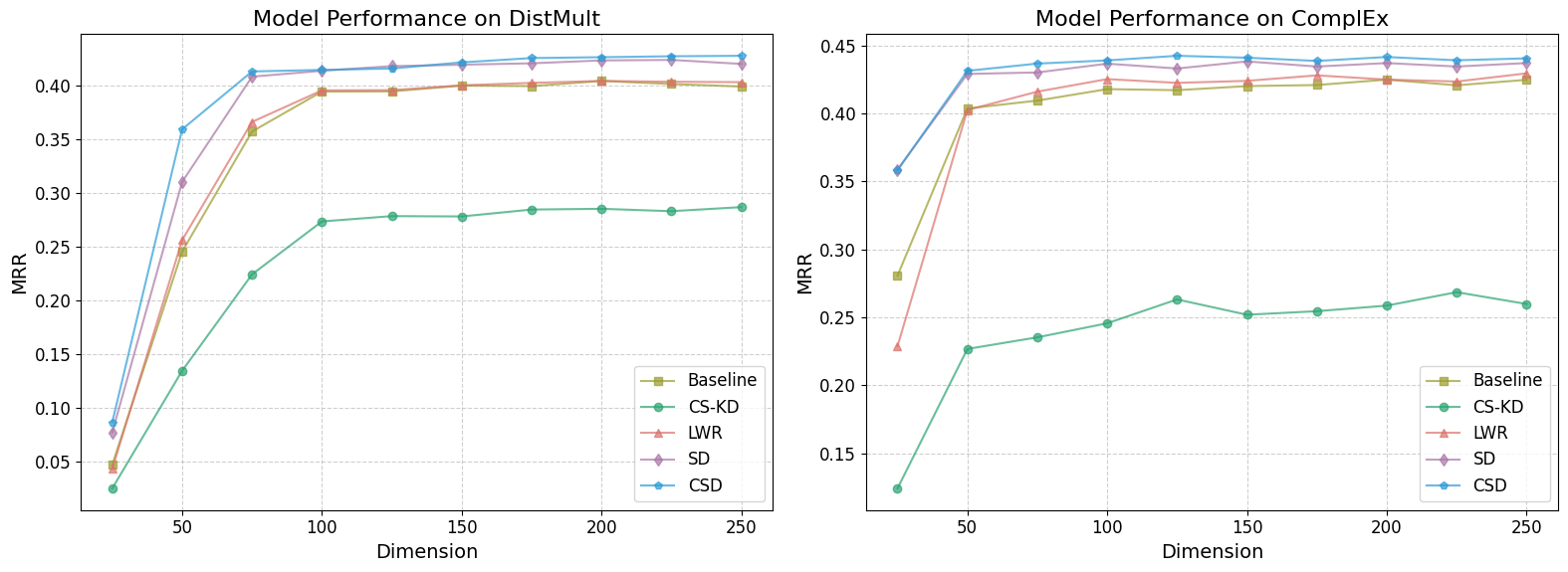}}
	\caption{The comparisons of link prediction performance (MRR) by DistMult and ComplEx on WN18RR with different embedding dimensions.}
	\label{dimension}
\end{figure*}

\begin{table*}[tbph]
	\setlength{\tabcolsep}{3mm}{}
	\centering
	\normalsize
	\caption{Ablation study: link prediction performance (MRR) on WN18RR and FB15K-237 with LowFER and TuckER with 25 dimensions.}\label{t4}
	\begin{tabular}[t]{ccccccc}
		\toprule
		& & Original & CSD (w/o $\mathcal{L}_{so}$) & CSD (w/o $\mathcal{L}_{se}$) & CSD (w/o Extractor) & CSD\\
		\midrule
		\multirow{3}*{WN18RR} & LowFER & .0160 & .0532 & .0305 & .0545 & \textbf{.0607} \\
		\cmidrule{2-7}
		& TuckER & .0644 & .1993 & .1840 & .1985 & \textbf{.2022}  \\
		\midrule
		\multirow{3}*{FB15K-237}& DistMult & .1983 & .2365 & .2309 & .2367 & \textbf{.2411} \\
		\cmidrule{2-7}
		& ComplEx & .2906 & .3074 & .3083 & .3067 & \textbf{.3113} \\
		\bottomrule
	\end{tabular}
\end{table*}

\begin{figure}[h]
	\centering
	\includegraphics[scale=0.42]{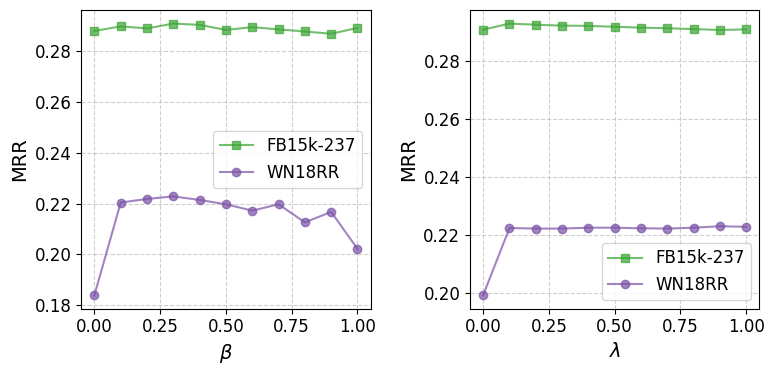}
        \caption{The Link prediction performance (MRR) with different value of hyperparameters $\beta$ and $\lambda$.}
	\label{fig:alpha_t}
\end{figure}

\subsection{Sensitivity Analyses}

\subsubsection{Different Embedding Dimensions. (RQ3)} We claim that our method can improve the expressiveness of KGE models for better link prediction in the low-dimension space. In section \ref{link prediction}, we experiment with the dimension setting 25 and 100. To comprehensively verify the performance of our CSD, we employ ComplEx and DistMult combined with our method to perform experiments on WN18RR with embedding dimensions ranging from 25 to 250 at intervals of 25. Other hyperparameters and experimental settings are identical to Section \ref{implement}. The performance results are shown in Figure \ref{dimension}. Generally speaking, all baselines combined with our CSD strategy outperform original models with all embedding dimensions. These positive findings indicate that the proposed CSD strategy is capable of boosting knowledge graph embedding model performance on the link prediction task with almost any embedding dimensions, especially the low embedding dimensions.

\subsubsection{Different value of hyperparameters. (RQ4)} In experiments, we apply the TuckER model to FB15k-237 and WN18RR datasets, both of which are of 25 dimensions. The hyperparameters $\beta$ and $\lambda$ are adjusted, incrementally from 0, by intervals of 0.1, up to $1.0$. The remaining settings are carried over from Section \ref{implement}. As depicted in Fig. \ref{fig:alpha_t}, there is an observable trend within the MRR that initially ascends and later descends with an increase in $\beta$ and $\lambda$. The maximum increment in MRR arises when $\beta$ and $\lambda$ rise from 0 to 0.1. Importantly, the MRR remains above the values found at $\beta=0$ or $\lambda=0$ across all other hyperparameter iterations, corroborating the efficacy of our proposed method. Further, the results suggest a dependence of the TuckER model's performance on both $\beta$ and $\lambda$ hyperparameters. Optimal values of these hyperparameters appear to be reliant on the specific dataset, emphasizing the necessity for conscientious tuning to achieve the best performance.

\subsubsection{Ablation Study (RQ5)}

To demonstrate the advantages of our confidence-aware self-semantic distillation, an ablation study is crucial. We performed these experiments using KGE models with the WN18RR and FB15K-237 dataset, adhering to the detailed parameter settings outlined in Section \ref{implement}. The results, as presented in Table \ref{t4}, signify that the methods without snapshot distillation (CSD(w/o $\mathcal{L}_{so}$)) and iterative self-semantic distillation (CSD(w/o $\mathcal{L}_{se}$)) surpass the performance of the original baselines. We also experimented with the method of directly using embeddings for distillation (CSD (w/o Extractor)). This method also surpasses the performance of the baselines. None of these methods surpass the performance of our CSD.

The model applying the CSD (w/o $\mathcal{L}_{so}$) method distills knowledge from the semantic extraction block and outperforms the initial baseline, which confirms that our method enhances the model's performance. The model applying the CSD (w/o $\mathcal{L}_{se}$) method training the model with some sub-processes, utilizing the snapshot of the model for knowledge distillation between two consequent sub-processes. Our approach, CSD, surpasses CSD (w/o $\mathcal{L}_{se}$) and CSD (w/o Extractor). This verifies the essential role of confidence-aware self-semantic distillation.

\section{Conclusions}\label{conclusions}
In this paper, we propose a confidence-aware self-semantic distillation strategy called CSD, which is model-agnostic and can be combined with any KGE model seamlessly. We design a novel semantic extraction block that extracts knowledge from embeddings and filters reliable knowledge by estimating the confidence of previously learned embeddings. The iteratively incorporated and accumulated iteration-based knowledge enhances the performance of the low-dimensional model, thereby improving the quality of link prediction in Knowledge Graphs (KGs). Experiments verify the effectiveness of our CSD. 

\bibliographystyle{plain}
\balance
\bibliography{sample-base}

\end{document}